\documentclass[conference]{IEEEtran}
\IEEEoverridecommandlockouts
\usepackage{cite}
\usepackage{amsmath,amssymb,amsfonts}
\usepackage{algorithmic}
\usepackage{graphicx}
\usepackage{textcomp}
\usepackage{xcolor}
\usepackage{float} 
\usepackage{capt-of}  
\usepackage{cuted}    
\usepackage{lipsum}
\usepackage{hyperref}

\def\BibTeX{{\rm B\kern-.05em{\sc i\kern-.025em b}\kern-.08em
    T\kern-.1667em\lower.7ex\hbox{E}\kern-.125emX}}
\begin{document}

\title{Vehicle Speed Detection System Utilizing YOLOv8: Enhancing Road Safety and Traffic Management for Metropolitan Areas \\
{\footnotesize \textsuperscript{}}

}

\author{

    \IEEEauthorblockN{SM Shaqib}
    \IEEEauthorblockA{
        \textit{Department of CSE}, \\
        \textit{Daffodil International University}, \\
        Dhaka, Bangladesh \\
        shaqib15-4614@diu.edu.bd}
    \and

    \IEEEauthorblockN{Alaya Parvin Alo}
    \IEEEauthorblockA{
        \textit{Department of CSE},\\
        \textit{Daffodil International University},\\ 
        Dhaka, Bangladesh \\
        alo15-4283@diu.edu.bd}
    \and

    \IEEEauthorblockN{Shahriar Sultan Ramit}
    \IEEEauthorblockA{
        \textit{Department of CSE},\\ 
        \textit{Daffodil International University},\\ 
        Dhaka, Bangladesh \\
        Shahriar15-4248@diu.edu.bd}
    \and

    \IEEEauthorblockN{Afraz Ul Haque Rupak}
    \IEEEauthorblockA{
        \textit{Department of CSE}, \\
        \textit{Daffodil International University}, \\
        Dhaka, Bangladesh \\
        afra15-13252@diu.edu.bd}
    \and

    \IEEEauthorblockN{Sadman Sadik Khan}
    \IEEEauthorblockA{
        \textit{Department of CSE}, \\
        \textit{Daffodil International University}, \\
        Dhaka, Bangladesh \\
        sadman15-13696@diu.edu.bd}
    \and
    
    \IEEEauthorblockN{Mr. Md. Sadekur Rahman}
    \IEEEauthorblockA{
        \textit{Department of CSE}, \\
        \textit{Daffodil International University},\\ 
        Dhaka, Bangladesh \\
        sadekur.cse@daffodilvarsity.edu.bd}

}

\maketitle

\begin{abstract}
In order to ensure traffic safety through a reduction in fatalities and accidents, vehicle speed detection is essential. Relentless driving practices are discouraged by the enforcement of speed restrictions, which are made possible by accurate monitoring of vehicle speeds. Road accidents remain one of the leading causes of death in Bangladesh. The Bangladesh Passenger Welfare Association stated in 2023 that 7,902 individuals lost their lives in traffic accidents during the course of the year. Efficient vehicle speed detection is essential to maintaining traffic safety. Reliable speed detection can also help gather important traffic data, which makes it easier to optimize traffic flow and provide safer road infrastructure. The YOLOv8 model can recognize and track cars in videos with greater speed and accuracy when trained under close supervision. By providing insights into the application of supervised learning in object identification for vehicle speed estimation and concentrating on the particular traffic conditions and safety concerns in Bangladesh, this work represents a noteworthy contribution to the area. The MAE was 3.5 and RMSE was 4.22 between the predicted speed of our model and the actual speed or the ground truth measured by the speedometer Promising increased efficiency and wider applicability in a variety of traffic conditions, the suggested solution offers a financially viable substitute for conventional approaches.

\end{abstract}


\section{Introduction}
Road accidents in Bangladesh have claimed a considerable number of lives throughout the last five years. Bangladesh Road Transport Authority (BRTA) and the Bangladesh Jatri Kalyan Samity, indicate that there are a significant number of fatal traffic accidents each year. Overcrowded roads, increased travel times, and worsening traffic accidents are the consequences of population growth and more cars [1]. These issues only become exacerbated by the insufficient infrastructural system that struggles to keep up with the increasing needs. Nowadays it is difficult to handle complex traffic systems by using conventional speed enforcement methods that are mostly dependent on humans and a few cheap and simple machines [4]. While also problematic due to coverage and accuracy reasons, these techniques cannot provide real-time data which is key to achieving optimal traffic management [5]. In that sense; smart solutions based on the latest technology are a must to reinforce the efficiency and reliability of traffic monitoring systems.
Deep learning and computer vision technologies [5] have allowed for the deployment of advanced traffic monitoring systems. YOLOv8, the most recent incarnation, brings significant improvements in speed, accuracy, and robustness, all while being suitable for real-time analysis applications [5]. The state of the art for object detection, YOLO models: YOLO (You Only Look Once) allows object detection from images or video streams with very high accuracy and (relatively) low latency [5], and has been a breakthrough in the area of object detection techniques available so far. As a consequence, a wide variety of applications (e.g., security monitoring, and autonomous driving [2]) are affected. Recent statistics show how dreadful traffic accidents are in Bangladesh. Of the total, 34.86 percent occurs on national highways, 28.41 percent on regional highways, and 28.5 percent on feeder roads [5]. In the Dhaka Metropolitan Area, 6.32 percent of all accidents occur; 1.11 percent of all accidents occur in the Chattogram Metropolitan Area [5]. Now the point is these alarming statistics demand the need for a sophisticated vehicle speed-detection system that can monitor and control traffic efficiently to reduce the number of accidents and improve road safety. 
Even though new tech can help, Bangladesh's traffic monitoring is old and not good [5].   The usual ways to check speed, like cameras and radar, have big problems [1].   They can't handle lots of data, make mistakes, and need people to check them a lot [1].  Also, they can't give real-time stats, which are very important to control traffic and stop crashes quickly [1]. 
These old ways don't work well. We really need a better system that can work alone and give good results fast [4]. Bangladesh's weird traffic, with lots of cars, many types of vehicles, and bad driving, makes it even harder for old systems to work [4]. The fix for this is to use the latest tech, like YOLOv8, with machine learning, to watch traffic [2].In short, using the YOLOv8 system can make roads safer in Bangladesh. It uses advanced technology to monitor and control traffic. This can improve how traffic is managed and make it better. This might help other places with similar traffic issues too. It can make roads safer worldwide.

\section{Literature Review}

Parwateeswar Gollapalli et al.[1] uses YOLOv8 for vehicle detection and Deep SORT for multi-object tracking to estimate the speeds of vehicles and count the number of vehicles in real-time. The Deep SORT model aims to provide high tracking accuracy to associate detected objects in consecutive video frames based on the features provided by Dense Re-ID. The system was tested with traffic of different levels of diversity and showed an accuracy level above 80\% of classifying vehicles and estimating their speed in a real dataset. This robust combination of YOLOv8 and Deep SORT is a dependable, low-cost method for high-speed traffic monitoring and management. Jin-xiang Wang [2] An algorithm for vehicle speed estimation by video surveillance The characteristic points of moving vehicles were extracted by using three–frame difference and background difference respectively. The position of vehicles is traced and their centroids are found so as to calculate their velocity by mapping the pixel distance to real distance. With a static camera (640x480, 18 fps), the combined methods achieved noise reduction and also detected boundaries accurately for the vehicles. The average speed estimation error of this approach is 0.097, which is robust enough to be applied to real-time traffic monitoring, even after the errors from the camera imaging principles. Genyuan Cheng et al.[3] offers a technique for video-based machine vision-based real-time vehicle speed detection. Using K-Nearest Neighbors (KNN) for target tracking and feature matching for background subtraction, the suggested method reduces algorithm complexity without sacrificing accuracy. Tests show that the approach performs well in real-time, with vehicle speed detection errors of about 5\%. Fatima Afifah et al.[4] suggests a technique that uses OpenCV and Python image processing to determine the speed of a vehicle. It attains high precision due to the Gaussian Mixture-based Background/Foreground Segmentation and contour detection. The 1920x1080 resolution and 24 frames per second video footage present in the database assure robust validation of the proposed method. Saif B.Neamah and Abdulamir A. Karim [5] present the work in the domain of real-time traffic monitoring systems using the YOLOv8 algorithm and other advanced deep learning for vehicle segmentation, classification, and detection. vehicle detection, tracking, speed estimation, and size estimation are the five stages of preprocessing. The system utilizes the Nvidia GTX 1070 GPU for exceptional precision in all the stages such
as 96.58\% for size estimation, 87.28\% for speed estimation, 97.54\%
for counting the vehicles, and 96.58\% for detecting and tracking vehicles. The evaluation dataset is made from 1920x1080 resolution and 24 frames per second video material. H´ector Rodr´ıguez-
Rangel et al.[6] With the objective of safety on roads without the use of specialized equipment, here is an advanced study to estimate the speeds of vehicles in real-time from a monocular camera
Data. The study evaluates statistical and machine learning approaches, like YOLOv3 and Kalman filtering, which are implemented on videos captured on highways. The LRM model is the most precise, and it demands the lowest processing power, it has competitive results and fantastic integration capabilities with Traffic Management and Urban Infrastructure. Jamuna S. Murthy et al. [7]the paper aims at road safety, and it presents a novel framework that uses the YOLOv5 object identification algorithm encountered in Advanced Driver Assistance Systems (ADAS).By incorporating real-time obstacle detection into the "ObjectDetect" mobile application, drivers may solve issues related to speed, accuracy, and cost by receiving timely alerts. The framework attempts to smoothly integrate ADAS into cars for increased driver safety. It does this by utilizing deep learning techniques, which demonstrate the progression from old methods. V. Barth et al. [8] presents an innovative way to improve traffic monitoring systems (TMS) by using a convolutional neural network (CNN) to recognize vehicles in video streams instead of the traditional Frame Subtraction method. The CNN-based approach maintains a low 5\% error rate for speed detection while achieving up to 12\% increase in vehicle detection accuracy, addressing constraints including changing lighting and pedestrian interference. This development is expected to result in safer and more effective transportation networks through more accurate and dependable traffic monitoring. Zhongmin Huangfu and Shuqing Li [9] present LW-YOLO v8, a lightweight target identification model designed for aerial photos taken by unmanned aerial vehicles (UAVs). Through the integration of lightweight convolution approaches and the Squeeze-and-Excitation (SE) module, the model improves the extraction of features from small targets while minimizing computing complexity. In comparison to YOLO v8n, experimental results on the VisDrone2019 dataset demonstrate a 3.8 percentage point gain in mean Average Precision (mAP@0.5) and a 7.2 GFLOP reduction in computing cost. An affordable option for precise small target identification in UAV aerial photos is provided by LW-YOLO v8. Yifan Shao et al.[10] suggests Aero-YOLO, a lightweight UAV image target identification technique built on top of YOLOv8, to overcome issues with tiny target sizes and restricted payload capacities. To improve computational efficiency and minimize model parameters, Aero-YOLO integrates the GSConv and C3 modules. Improved feature extraction results in faster and more accurate vehicle and pedestrian identification under a variety of situations. This is achieved through the use of the CoordAtt and Shuffle attention processes. Qi Liu et al. [11]Precise target recognition is still critical to improving traffic management effectiveness and road safety in the context of intelligent traffic systems. YOLOv8-SnakeVision is our suggested answer to persistent problems like accurate multi-object identification and complex traffic scenarios. It combines the Wise-IoU approach, Context Aggregation Attention Mechanisms, and Dynamic Snake Convolution within the YOLOv8 framework. This novel method provides strong assistance for the development of intelligent traffic systems by greatly improving small item detection and the recognition of hidden objects. Jozef Gerát et al.[12] suggests a system that uses video records to determine the speed of a vehicle. It makes use of techniques including optical flow, DBSCAN, Kalman filter, and Gaussian mixture models. Important elements include object detection and tracking, where speed is estimated using pixel movement rates. Testing shows how well the system works at various speeds and in various environments. In traffic monitoring systems, this software-based method provides an alternative to radar-based speed measurement. Uddeshya Gupta et al.[13]provides an affordable substitute for radar-based techniques for vehicle speed detection by putting forth a video-based solution. It makes use of tracking, object detection, speed estimation, and real-time video capture. With a high detection and measurement accuracy of more than 95\% for vehicle detection and more than 92\% for speed measurement, the system seeks to improve traffic management and road safety.Shuai Hua et al.[14] offers a novel method for estimating car speeds from traffic surveillance footage, tackling problems with urban traffic control. The paper suggests a solution for the 2018 NVIDIA AI City Challenge Track 1, focusing on effective speed estimation through vehicle identification and tracking, by combining deep learning and computer vision techniques. The model obtains promising results by utilizing optical flow for trajectory analysis and transfer learning for detection, thereby advancing the field of intelligent transportation systems. Alexander Grents et al.[15]presents a system that estimates traffic flows and calculates vehicle speed using data from video surveillance. For vehicle recognition and tracking, it makes use of cutting-edge methods as the SORT tracker and Faster R-CNN detector. Vehicle classification is accurate when a Mask R-CNN neural network trained on a big dataset is used. The technology determines vehicle speed by converting image coordinates to real-world scales, which has the potential to improve road safety and efficiency. Diogo Carbonera Luvizon et al.[16] presents an unobtrusive video-based technology intended to monitor vehicle speed accurately on city streets. Our method locates car license plates in motion-captured picture regions with high efficiency by utilizing an innovative text detector and an enhanced motion detector. Next, unique characteristics are identified, followed between frames, and used to calculate vehicle speed by comparing the trajectories to established real-world measurements. The system's effectiveness is demonstrated through experimental evaluation on a variety of datasets. It achieves speeds with an average inaccuracy of -0.5 km/h and outperforms other state-of-the-art text detectors in terms of recall and precision. Mohit Chandorkar et al. [17] explains a computer vision technique for tracking speed and detecting vehicles. Through the explanation of procedures like background subtraction and feature extraction, it emphasizes the significance of these systems for traffic control and safety. Different vehicle identification techniques are covered, along with a snippet of a Python implementation for OpenCV-based speed estimation and automobile tracking. The method's overall goal is to improve traffic safety and surveillance by utilizing computer vision and image processing techniques. Shengnan Lu1 et al. [18] presents a novel approach to vehicle speed estimation from video sequences. The method accomplishes accurate and real-time speed estimates by avoiding feature extraction problems and utilizing projection histograms in conjunction with a frame difference approach. With an average error of 0.3 km/h and 99.4\% of estimations falling within a 2 km/h error range, the experimental findings demonstrate good performance, making it appropriate for real-world applications. Hakan Koyuncu and  Baki Koyuncu et al.[19]suggests the use of a camera and image processing to identify the speed of a vehicle. For precise speed calculations below 50 km/h, two methods are used: discrete motion detection and linear motion detection. The system uses edge detection, background subtraction, and noise reduction as image processing processes. 
Distances are estimated using trigonometry, and speed is computed by timing how long it takes to pass a predetermined test area. The system's efficacy for real-time speed monitoring is demonstrated by the results, which show near alignment with automobile speedometers. JunmeiGuo et al. [20], on the other hand, focus on the suitability of the YOLOv8 model for the detection of foreign object intrusion on the highway, with a focus on the development of deep learning-based object recognition and tracking technologies.

\section{Methodology}
YOLOv8 employs several processes to estimate a vehicle's speed, most of which rely on its powerful object detection and categorization capabilities. The general principle behind these steps is that video content is split into individual frames, with these serving as the input to YOLOv8. Once the frames have been extracted, YOLOv8 applies its powerful neural network model architecture to each frame to adequately detect and categorize automobiles by identifying bounding boxes around every vehicle. These bounding boxes can be utilized to monitor the movement of each vehicle with a serial number over a series of frames. The change in the vehicle's position between frames is used to calculate its speed, taking into consideration the interval between these frames, which is determined by the frame rate of the video. This process leverages YOLOv8's high object detection accuracy to provide a reliable real-time vehicle speed measurement.
\newpage

\begin{strip} 
\includegraphics[width=1\linewidth,height=0.35\textheight]{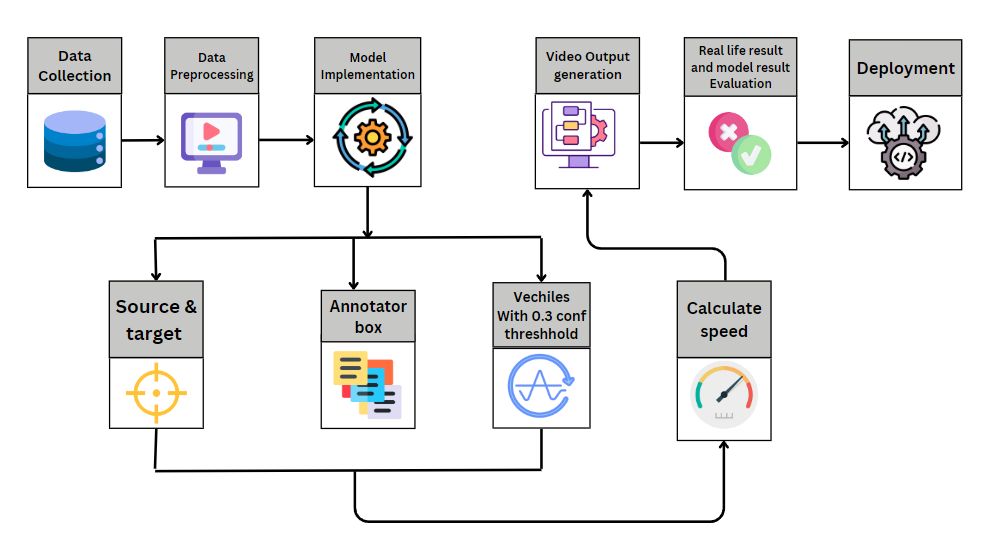}
\captionof{figure}{Porposed Methodology of Speed Detection System}
\label{Figure:1}
\end{strip}

\subsection{YOLOv8 Architecture}\label{AA}
The latest version of the You Only Look Once series: YOLOv8, it is at object detection and classification images quickly and accurately. It builds on the previous iterations by utilizing the latest techniques to optimize their missing.

\textbf{Core Network: }
YOLOv8 bases its feature extraction on CSPDarknet53. Cross-Stage Partial (CSP) connections are a
part of this architecture, enhancing gradient stability and information flow through training.

\textbf{Head and Neck Structures: }
The Path Aggregation Network (PANet) that constitutes the neck structure of YOLOv8 helps to collect
features at multiscale by enabling the flow of information through multiple spatial resolutions. The head is constructed by multiple detecting heads, each predicting the bounding boxes, class
probabilities, and objectness scores on different scales.

\textbf{Detection Head: }
The detecting head of YOLOv8 is another novel device containing a new IoU (Intersection over Union) loss function and dynamic anchor assignment.

\begin{figure}
    \centering
    \includegraphics[width=1\linewidth,height=0.3\textheight]{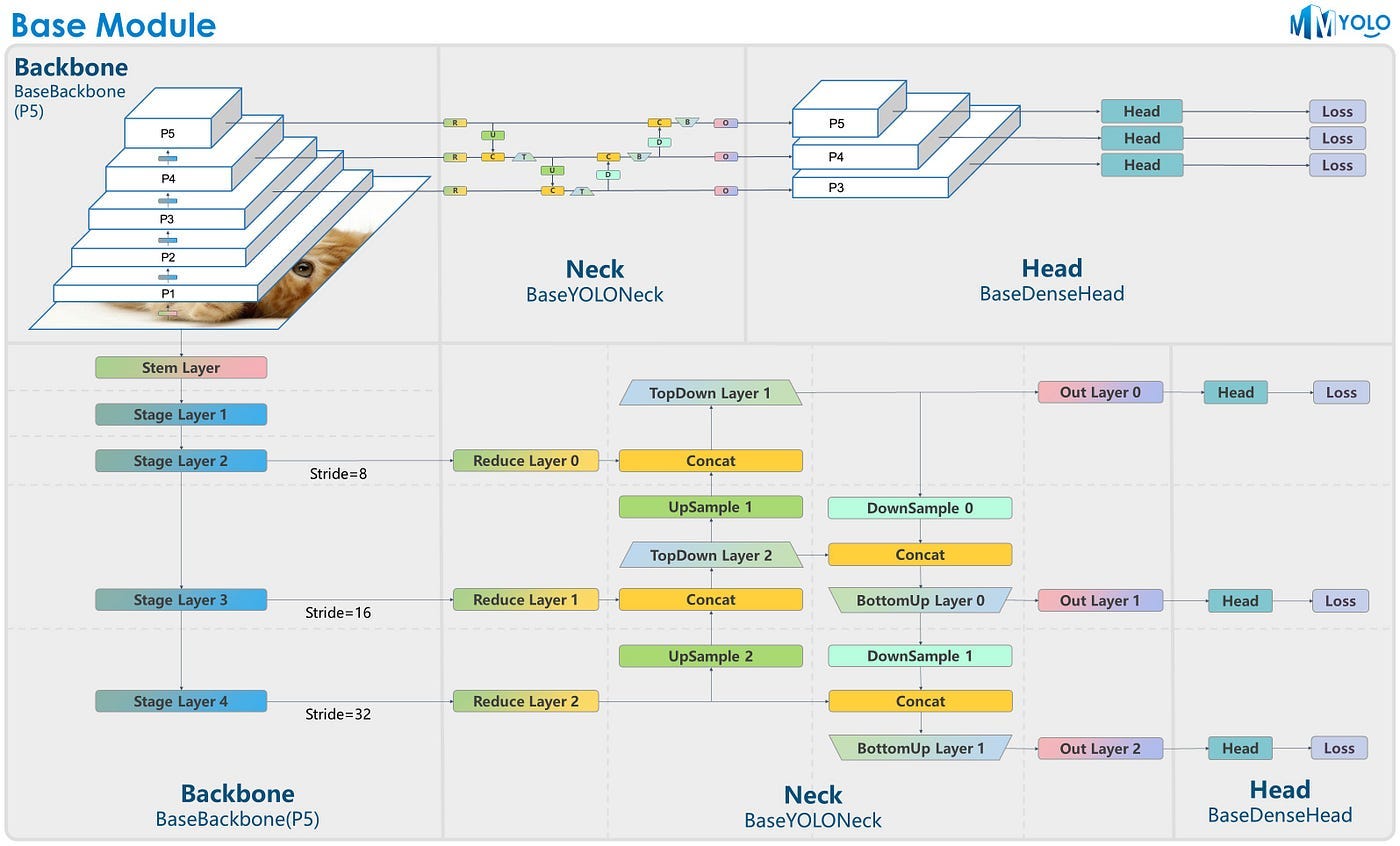}
    \caption{YOLOv8 Network Architecture}
    \label{fig:2}
\end{figure}

\subsection{Dataset collection \& Preprocessing}\label{AA}
The first step in the procedure is data gathering, which involves utilizing a camera to record video footage from a route. Following the use of a camera to gather video footage from the road, the video must be processed by being divided into multiple segments or spots. There are several reasons to segment the data, including narrowing down on particular stretches of the road, lightening the computational burden during analysis, or identifying pertinent occurrences of vehicle movement.

\subsection{Determining Source and Target ROIs}

The implementation of vehicle detection and tracking algorithms within this designated zone can begin once the source, target, and ROI have been established. Vehicles within the ROI can be found using YOLOv8's object detection capabilities.\

\textbf{Source and Target Identification: }
A vehicle's beginning point of motion inside the frame is referred to as its source. When the car first enters the camera's frame of view, it is located in this position. The target, as it appears within the frame, is the destination or last position of the vehicle's motion. This is the point at which the car leaves the frame of the camera or arrives at a certain location.

\textbf{Region of Interest (ROI): }
The ROI is a designated region in the video frame where the movement of the vehicles will be tracked and examined. Based on the particular route or area of interest that is visible within the camera's field of view, this area was chosen.
The regions of interest (ROI) that are relevant for vehicle speed detection include traffic lanes, intersections, and other places. Reducing computational overhead and increasing accuracy, the ROI is defined, and vehicle identification and tracking algorithms are focused on the pertinent area. The Region of Interest of ROI is shown in Figure 3 and Figure 4 with different videos of roads.

\begin{figure}
    \centering
    \includegraphics[width=1\linewidth,height=0.25\textheight]{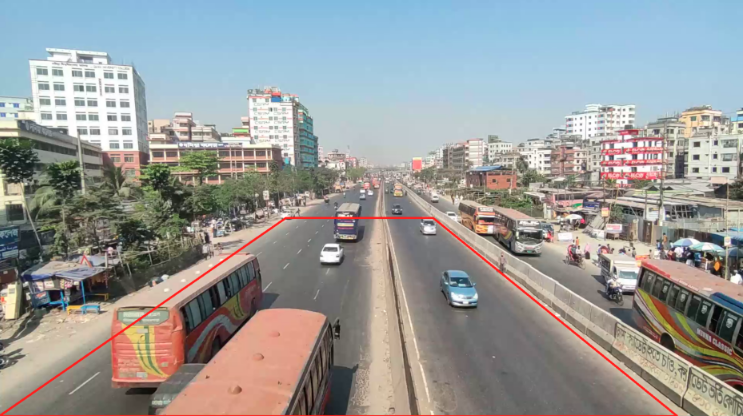}
    \caption{Region of Interest ROIs for the first road}
    \label{fig:2}
\end{figure}

\begin{figure}
    \centering
    \includegraphics[width=1\linewidth,height=0.25\textheight]{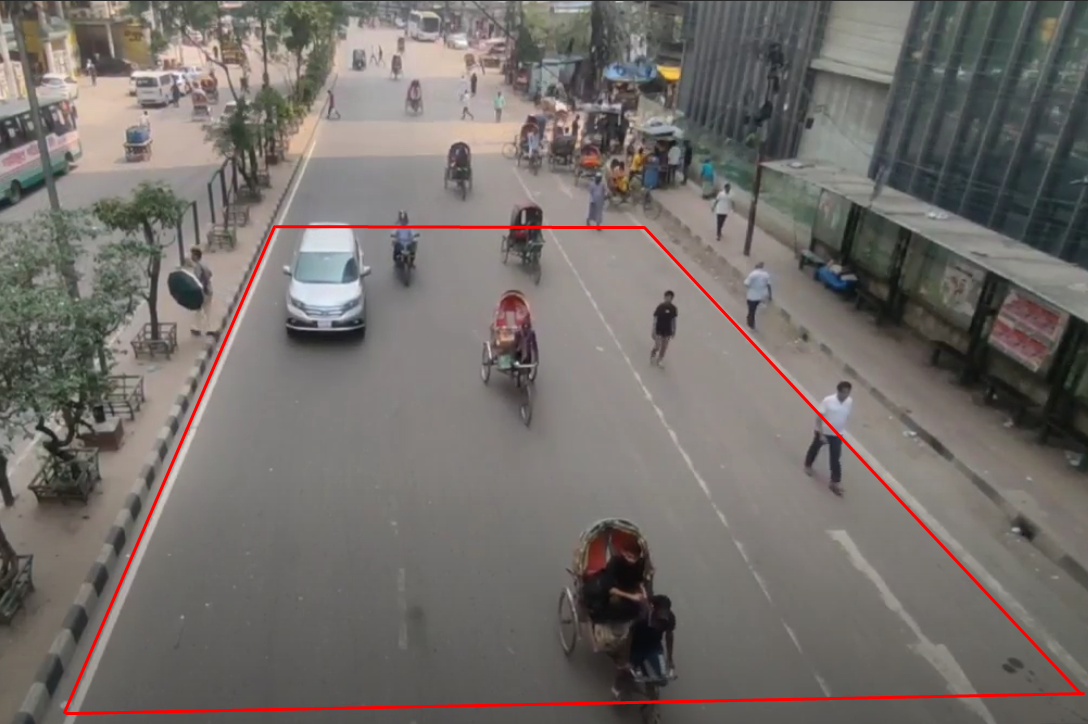}
    \caption{Region of Interest ROIs for the second road}
    \label{fig:2}
\end{figure}

\subsection{Transform Perspective}
The transformation matrix M, which transfers points from the source plane (original image) to the target plane (desired perspective), must be located to accomplish the perspective transformation. Finding a matrix that precisely translates points from one plane to another is necessary for the transformation matrix calculation, which allows for perspective correction and alignment of regions of interest in computer vision tasks. Accurate spatial relationships between objects are critical for tasks like object tracking, panorama stitching, and image rectification, for which this transformation is required.

\begin{figure}
    \centering
    \includegraphics[width=1.1\linewidth,height=0.25\textheight]{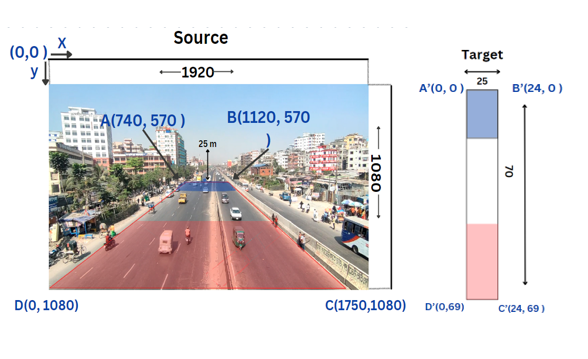}
    \caption{Source and Target Identification}
    \label{fig:2}
\end{figure}

\textbf{a) Source and Target Points: }
We start at four matching points in the source and target planes. we can see the source and target region in the figure 5.\\

Source Points:\(A(x_1, y_1), B(x_2, y_2), C(x_3, y_3), D(x_4, y_4)\)

Target Points:  \(A(u_1, v_1), B(u_2, v_2), C(u_3, v_3), D(u_4, v_4)\).

\textbf{b) Homogeneous Coordinates: }
We represent these points in homogeneous coordinates by appending a 1 to each point

Source point : \((x_1, y_1, 1), (x_2, y_2, 1), (x_3, y_3, 1), (x_4, y_4, 1)\).

Target point : \((u_1, v_1, 1), (u_2, v_2, 1), (u_3, v_3, 1), (u_4, v_4, 1)\).

\textbf{c)  Homography Matrix: }
The source and target homogeneous points can be used to calculate the transformation matrix M, also referred to as the homography matrix. Usually, the least-squares approach or the Direct Linear Transformation (DLT) technique are used for this.

\textbf{d) Applying Transformation }
Once we have the transformation matrix 
M, we can apply it to transform the coordinates of all points in the source plane to their corresponding points in the target plane.

\textbf{e) Matrix Multiplication: }
Given a point $\begin{pmatrix} x_s, y_s , 1 \end{pmatrix}$ in the source plane represented in homogeneous coordinates, we perform matrix multiplication with $M$ to obtain the transformed point $\begin{pmatrix} x_t, y_t , w \end{pmatrix}$ in the target plane:\\

$\begin{pmatrix} x_t \\ y_t \\ 1 \end{pmatrix} = M \cdot \begin{pmatrix} x_s \\ y_s \\ 1 \end{pmatrix}$\\

\textbf{f) Normalization: }
To obtain the final transformed coordinates $\begin{pmatrix} x_t , y_t \end{pmatrix}$
in the non-homogeneous form, we normalize by dividing the transformed homogeneous coordinates by the third component w.

\subsection{Speed Calculation}
The provided code calculates speed by gathering positioning data from moving vehicles over time, calculating time and distance traveled, and then translating the result to the chosen speed unit. Real-time assessment of the vehicle speeds in the video is made possible by this approach.

\textbf{a) Distance Calculation: }
The absolute difference between the vehicle's beginning and finishing positions within the designated Region of Interest (ROI) is used to compute the vehicle's distance traveled. During the tracking period, this distance indicates the vehicle's displacement within the ROI.

\textbf{b) Time Calculation: }
The time elapsed is determined based on the number of frames over which the positional data is collected and the frame rate (FPS) of the video. We denoted the number of frames as n and the frame rate as FPS. The total time taken for n frames is calculated using the formula $\text{Total Time} = \frac{n}{\text{FPS}}$. If the positional data is collected over n frames, the time taken for each frame is $\frac{1}{\text{FPS}}$. 

\textbf{c) Speed Calculation: }
The speed of a vehicle is calculated using the formula:\\ $\text{Speed} = \frac{\text{Time}}{\text{Distance}} \times \text{Conversion Factor.}$ \\
Once the distance and time are determined, the speed is calculated using the formula. The distance is divided by the time to obtain the speed in units of distance per time (pixels per frame). To convert the speed to a specific unit (kilometers per hour), a conversion factor is applied

\subsection{Evaluation of Model}
Mean Absolute Error (MAE) and Root Mean Square Error (RMSE) are two important metrics I utilized to assess the efficacy of my vehicle speed detection model. Using both MAE and RMSE, I was able to get a thorough assessment of the accuracy of my model. While RMSE showed how higher errors affected model performance, MAE gave a clear picture of the average prediction error. All things considered, these measurements showed that my vehicle speed detection algorithm operates fairly accurately, despite the fact that some larger errors continue to significantly affect the error statistic as a whole.\\

\textbf{MAE: } Mean Absolute Error (MAE) is a statistical measure that quantifies the average magnitude of errors in a set of predictions, without considering their direction. It is calculated as the average of the absolute differences between predicted values and actual values. MAE is used to evaluate the accuracy of a predictive model by measuring how close the predicted values are to the actual outcomes.

$\text{MAE} = \frac{1}{n} \sum_{i=1}^{n} \left| \text{Predicted}_i - \text{Actual}_i \right|$\\

\textbf{RMSE: } The average amount of errors in a group of predictions is quantified by the statistical measure known as Root Mean Square Error (RMSE). The square root of the average of the squared discrepancies between the expected and actual values is computed. Relative mean square error (RMSE) is a frequently used metric to assess the accuracy of predictive models. It is especially useful in regression analysis since it indicates how well the model predicts the real data.\\

$\text{RMSE} = \sqrt{\frac{1}{n} \sum_{i=1}^{n} \left( \text{Predicted}_i - \text{Actual}_i \right)^2}$ \newpage

\section{Result \& Discussion}

As illustrated in Figure 6, the video output is produced by the suggested approach. This output provides precise speed calculations and assigns a unique ID number to each vehicle. Vehicles are detected using bounding boxes, and accurate tracking and identification are ensured by setting the confidence threshold at 0.3.
The comparison involves analyzing the deviations of the ground truth measurements and the model's speed estimates. The difference between the speedometer readings and the model's estimates can be calculated to quantify the model's output accuracy using measures like mean absolute error (MAE) or root mean square error (RMSE). The model's speed estimates are then validated based on the level of concurrence with the exact measurements on the ground. High agreement of the model outputs with the speedometer readings will imply the high accuracy of the speed detection method. Figure 8 shows the graph and comparison of RMSE vs MAE value and Predicted values vs Actual value

\begin{figure}[H]
    \centering
    \includegraphics[width=1\linewidth,height=0.26\textheight]{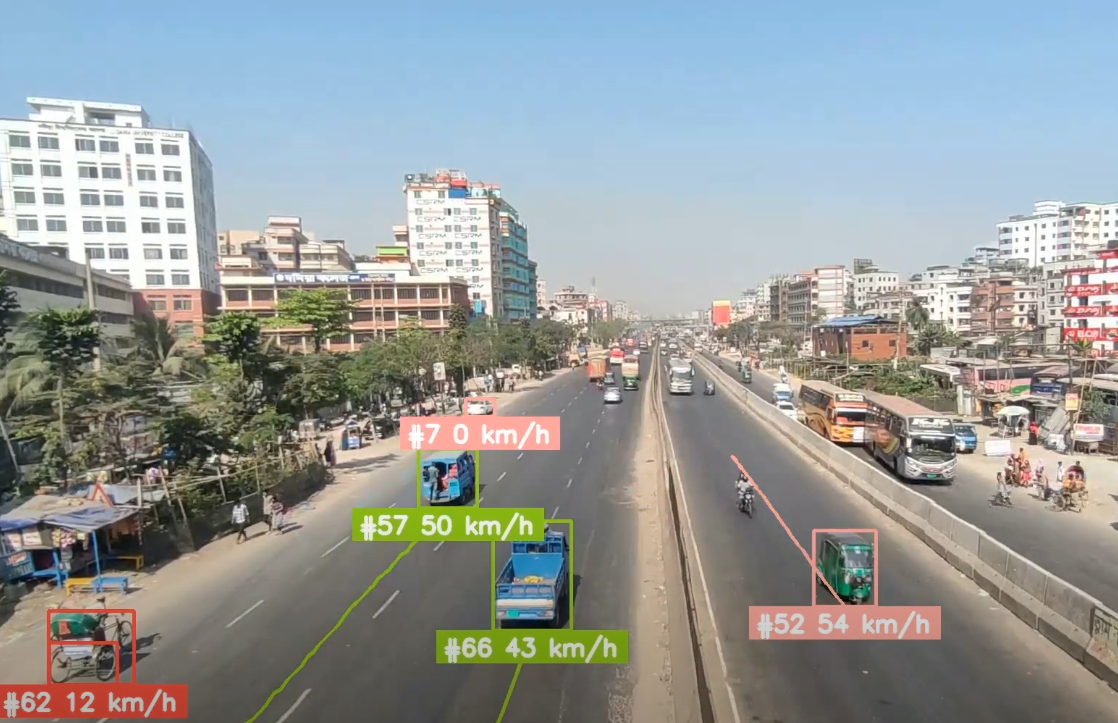}
    \caption{Video output of Speed Measurement}
    \label{fig:2}
\end{figure}

\begin{figure}[H]
    \centering
    \includegraphics[width=1\linewidth,height=0.26\textheight]{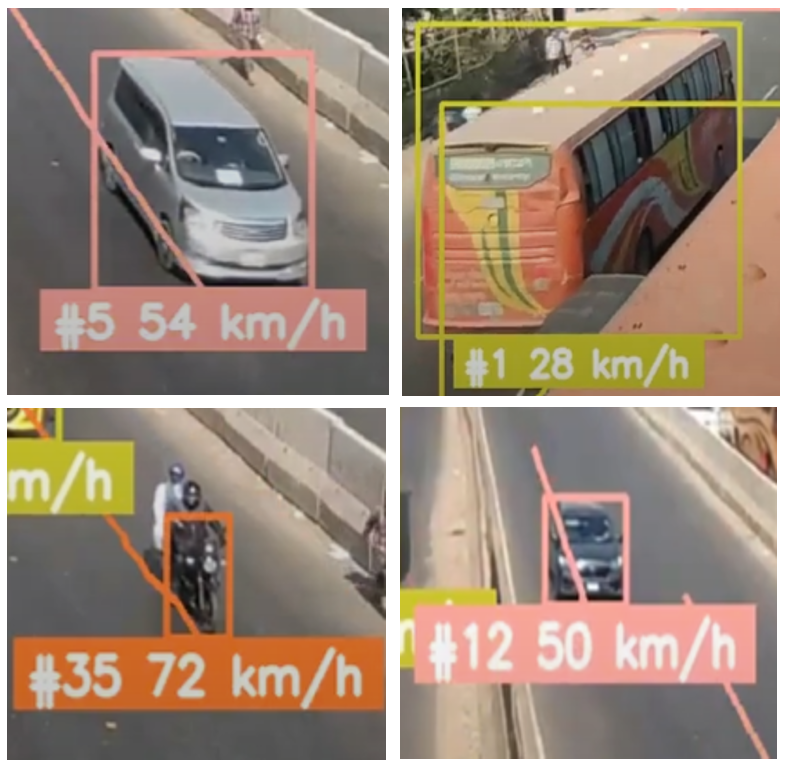}
    \caption{Speed Tracking of Vehicles by Unique ID}
    \label{fig:2}
\end{figure}

\newpage

\begin{table}[h]
\centering
\caption{Comparison of Predicted and Actual Values}
\renewcommand{\arraystretch}{1.2} 
\begin{tabular}{|c|c|c|c|c|}

\hline
\text{Car} & \text{Predicted} & \text{Actual} & $(\text{P} - \text{A})^2$ & $|\text{P} - \text{A}|$ \\

\text{Id} & \text{(P)} & \text{(A)} &  &  \\

\hline
1 & 29 & 25 & $(29-25)^2 = 16$ & $|29-25| = 4$ \\
2& 21 & 20 & $(21-20)^2 = 1$ & $|21-20| = 1$ \\
3&58 & 55 & $(58-55)^2 = 9$ & $|58-55| = 3$ \\
4& 50 & 54 & $(50-54)^2 = 16$ & $|50-54| = 4$ \\
5& 56 & 50 & $(56-50)^2 = 36$ & $|56-50| = 6$ \\
6 & 57 & 54 & $(57-54)^2 = 9$ & $|57-54| = 3$ \\
7 & 0 & 0 & $(0-0)^2 = 0$ & $|0-0| = 0$ \\
30 & 36 & 33 & $(36-33)^2 = 9$ & $|36-33| = 3$ \\
31 & 57 & 54 & $(57-54)^2 = 9$ & $|57-54| = 3$ \\
35&  68 & 66 & $(68-66)^2 = 4$ & $|68-66| = 2$ \\
20 & 16 & 12 & $(16-12)^2 = 16$ & $|16-12| = 4$ \\
12 & 50 & 50 & $(50-50)^2 = 0$ & $|50-50| = 0$ \\
18 & 32 & 30 & $(32-30)^2 = 4$ & $|32-30| = 2$ \\
42 & 57 & 53 & $(57-53)^2 = 16$ & $|57-53| = 4$ \\
41 & 50 & 56 & $(50-56)^2 = 36$ & $|50-56| = 6$ \\
54 & 56 & 58 & $(56-58)^2 = 4$ & $|56-58| = 2$ \\
57 & 54 & 56 & $(54-56)^2 = 4$ & $|54-56| = 2$ \\
66 & 43 & 46 & $(43-46)^2 = 9$ & $|43-46| = 3$ \\
62 & 17 & 12 & $(17-12)^2 = 25$ & $|17-12| = 5$ \\
73 & 61 & 57 & $(61-57)^2 = 16$ & $|61-57| = 4$ \\
77 & 18 & 13 & $(18-13)^2 = 25$ & $|18-13| = 5$ \\
84 & 72 & 66 & $(72-66)^2 = 36$ & $|72-66| = 6$ \\
89 & 56 & 48 & $(56-48)^2 = 64$ & $|56-48| = 8$ \\
96 & 36 & 30 & $(36-30)^2 = 36$ & $|36-30| = 6$ \\
62 & 21 & 17 & $(21-17)^2 = 16$ & $|21-17| = 4$ \\
83 & 57 & 50 & $(57-50)^2 = 49$ & $|57-50| = 7$ \\
92 & 54 & 49 & $(54-49)^2 = 25$ & $|54-49| = 5$ \\
100 & 10 & 12 & $(10-12)^2 = 4$ & $|10-12| = 2$ \\
98 & 54 & 49 & $(54-49)^2 = 25$ & $|54-49| = 5$ \\
109 & 28 & 21 & $(28-21)^2 = 49$ & $|28-21| = 7$ \\
132 & 50 & 47 & $(50-47)^2 = 9$ & $|50-47| = 3$ \\
143 & 57 & 50 & $(57-50)^2 = 49$ & $|57-50| = 7$ \\
145 & 25 & 21 & $(25-21)^2 = 16$ & $|25-21| = 4$ \\
149 & 54 & 51 & $(54-51)^2 = 9$ & $|54-51| = 3$ \\
168 & 61 & 55 & $(61-55)^2 = 36$ & $|61-55| = 6$ \\
176 & 31 & 28 & $(31-28)^2 = 9$ & $|31-28| = 3$ \\
145 & 0 & 0 & $(0-0)^2 = 0$ & $|0-0| = 0$ \\
186 & 43 & 40 & $(43-40)^2 = 9$ & $|43-40| = 3$ \\
193 & 54 & 50 & $(54-50)^2 = 16$ & $|54-50| = 4$ \\
224 & 46 & 43 & $(46-43)^2 = 9$ & $|46-43| = 3$ \\
\hline
 Total   &  1695  &  1581  &      712   &   140  \\
 \hline

\end{tabular}

\label{tab:prediction_comparison}
\end{table}

\textbf{Sum of Squared Errors (SSE): } 712


\textbf{Sum of Absolute Errors (SAE): } 140\\


After calculating SSE and SAE we can calculate the value of RMSE and MAE.\\

\textbf{Number of observations:  } (n) = 40

RMSE = $\sqrt{\frac{\sum_{i=1}^{n}(Predicted_i - Actual_i)^2}{n}} = \sqrt{\frac{712}{40}} \\            {\approx 4.22}$\\

\large
MAE = $\frac{\sum_{i=1}^{n}|Predicted_i - Actual_i|}{n} = \frac{140}{40} = 3.5$\\

The average error between the actual speed and predicted speed of the model is calculated with the formula of percentage error 
\[
\text{Percentage Error} = \left| \left(\frac{\sum \text{Predicted}}{\sum \text{Actual}}\right) - 1 \right| \times 100
\]

\[
\text{Percentage Error} = \left| \left(\frac{\sum \text{1695}}{\sum \text{1581}}\right) - 1 \right| \times 100
\]

The accuracy of the model is 92.79\% for the speed calculated by the speedometer and predicted speed value.

\begin{figure}
    \centering
    \includegraphics[width=1\linewidth,height=0.35\textheight]{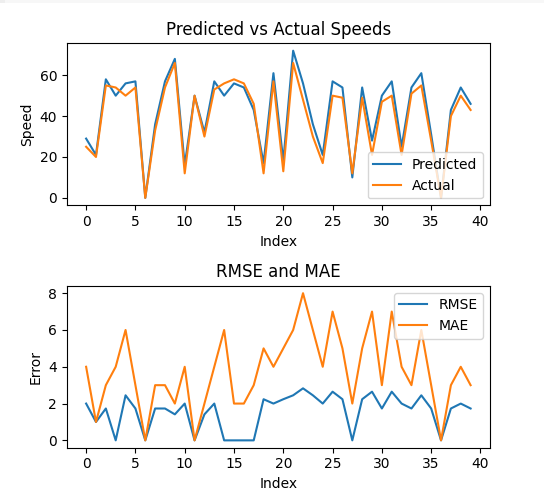}
    \caption{The Comparison of the Predicted vs actual speed and the RMSE vs MAE}
    \label{fig:2}
\end{figure}

\section{Conclusion \&Future Work}

Traffic accidents are a growing concern in Bangladesh, particularly on highways and roads like those in Gulshan, with nighttime incidents being especially frequent. Our research aims to develop an automated alert system to notify police immediately in the event of an accident. This system will enhance human safety by ensuring quick response times, ultimately helping to reduce the impact of road accidents. In the present research, an advanced vehicle speed detection system is developed using the YOLOv8 model, and its application in improving road safety and traffic management in Bangladesh is described. YOLOv8, with the ability of deep learning, accurately detects and traces vehicles in real-time for their speed enforcement. Our method presented better performance under varying traffic scenarios, which is critical for the reduction of accidents and fatal injuries caused on the roads. Furthermore, it is cost-effective and can be easily scaled up. This study highlights that advanced deep learning techniques are crucial for better traffic safety and traffic management. Further research could focus on model fine-tuning and testing in real situations in multiple parts of the world. 

\rule{0.5\textwidth}{0.4pt}

Github link for code:\url{https://github.com/smshaqib/Research/blob/main/Vechile%20speed/VECHILE_SPEED_DETECTION_SYSTEM.ipynb}.

\end{document}